\documentclass[10pt, format=sigconf,nonacm]{acmart}
\renewcommand\footnotetextcopyrightpermission[1]{}
\AtBeginDocument{%
  \providecommand\BibTeX{{%
    \normalfont B\kern-0.5em{\scshape i\kern-0.25em b}\kern-0.8em\TeX}}}

\usepackage[capitalize]{cleveref}
\crefname{section}{Sec.}{Secs.}
\Crefname{section}{Section}{Sections}
\Crefname{table}{Table}{Tables}

\usepackage[ruled,linesnumbered]{algorithm2e}
\usepackage{fancyhdr}
\usepackage[misc]{ifsym} 

\begin{document}

\title{Supervised Contrastive Learning with Structure Inference for Graph Classification}


\author{Hao Jia}
\affiliation{%
  \institution{Beijing University of Technology}
  \city{Beijing}
  \country{China}}

\author{Junzhong Ji}
\affiliation{%
  \institution{Beijing University of Technology}
  \city{Beijing}
  \country{China}}

\author{Minglong Lei \textrm{*}}
\affiliation{%
  \institution{Beijing University of Technology}
  \city{Beijing}
  \country{China}}
\email{leiml@bjut.edu.cn}

\begin{abstract}
Advanced graph neural networks have shown great potentials in graph classification tasks recently. Different from node classification where node embeddings aggregated from local neighbors can be directly used to learn node labels, graph classification requires a hierarchical accumulation of different levels of topological information to generate discriminative graph embeddings. Still, how to fully explore graph structures and formulate an effective graph classification pipeline remains rudimentary. In this paper, we propose a novel graph neural network based on supervised contrastive learning with structure inference for graph classification. First, we propose a data-driven graph augmentation strategy that can discover additional connections to enhance the existing edge set. Concretely, we resort to a structure inference stage based on diffusion cascades to recover possible connections with high node similarities. Second, to improve the contrastive power of graph neural networks, we propose to use a supervised contrastive loss for graph classification. With the integration of label information, the one-vs-many contrastive learning can be extended to a many-vs-many setting, so that the graph-level embeddings with higher topological similarities will be pulled closer. The supervised contrastive loss and structure inference can be naturally incorporated within the hierarchical graph neural networks where the topological patterns can be fully explored to produce discriminative graph embeddings. Experiment results show the effectiveness of the proposed method compared with recent state-of-the-art methods.
\end{abstract}


\maketitle

\section{Introduction}
\label{sec:intro}
Graphs are universal data structures that can model many real-world systems with complex interactions, e.g., social networks, biological networks, and traffic networks. Graph classification is a fundamental pattern recognition task that can largely improve the understanding of the functions of those systems. The goal of graph classification is to summarize the patterns in topological structures and node attributes to train a competitive model that can classify each graph into a category \cite{zhang2018end}. Successfully resolving this problem provides important support for many downstream applications, e.g., molecular property prediction \cite{gilmer2017neural,shui2020heterogeneous}, brain disease classification \cite{parisot2018disease,li2021braingnn}, point clouds classification \cite{wang2019dynamic,nezhadarya2020adaptive}, and text classification \cite{yao2019graph,liu2020tensor}.

Early methods that resolve graph classification tasks are mainly based on graph kernels that define functions to measure the similarities between different graphs \cite{vishwanathan2010graph, shervashidze2011weisfeiler}. With the structural similarities captured by graph kernels, the kernelized learning method, e.g., Support Vector Machines (SVMs), can be leveraged to perform the classification \cite{yanardag2015deep}. General options for graph kernels include random walk kernels \cite{sugiyama2015halting}, shortest-path kernels \cite{borgwardt2005shortest} and Weisfeiler-Lehman kernels \cite{shervashidze2011weisfeiler}. However, these methods highly depend on handcrafted kernels to extract structural patterns in graphs without explicit graph representations in most cases, which largely limits their integration with general classification models \cite{narayanan2017graph2vec}.

Recently, graph neural networks (GNNs) has been widely used in graph classification tasks. In general, GNNs use a local aggregation function to update node features by iteratively aggregating information from neighbors, and then employ a global readout function to summarize the node features as graph embeddings. The classification results can be easily obtained by applying an additional classification layer. From the local perspective, the information aggregations through edges are correlated with the feature forward process in different types of neural networks, e.g., MLP \cite{hamilton2017inductive}, attention networks \cite{velivckovic2018graph}, and jumping connections \cite{xu2018representation}. From the global perspective, the acquisition of graph embeddings depends on pooling methods which progressively generate coarser graphs to capture hierarchical structures \cite{ma2019graph, ying2018hierarchical,li2020graph}. With carefully designed aggregation and readout functions, GNNs are as powerful as the Weisfeiler-Lehman Isomorphism test \cite{xu2018powerful}.

Despite much progress of graph neural networks, thoroughly exploring the hierarchical structures and learning generalizable graph-level features for classification remains a challenge. To go beyond their limitations, recent methods use contrastive self-supervised learning to improve the feature extraction ability of GNNs \cite{velickovic2019deep, peng2020graph, hassani2020contrastive, sun2021sugar}. A common approach is using graph augmentation to develop different views for graphs so that the contrastive objective can be easily constructed \cite{you2020graph, zhu2021graph}. However, on the one hand, the augmentation strategies in these methods are usually defined from underlying priors, e.g., node dropping, attribute masking. Such formulation may bring about model biases toward specific graph structures or tasks. Besides, some of those methods are designed for node-level classifications, which may encounter difficulties when applied to graph classifications. On the other hand, the widely used contrastive learning frameworks for graphs are usually established in a self-supervised manner. The contrastive losses in those methods are based on one-positive versus one-negative or many-negatives paradigms where the contrastive power is limited.

In this paper, we address above problems by proposing a novel graph classification method that incorporates \textbf{Sup}ervised \textbf{co}ntrastive learning with \textbf{s}tructure \textbf{in}ferenc\textbf{e} into graph neural networks   (\textbf{SupCosine}). The goal is to fully utilize the graph structures and label information to formulate an effective graph contrastive learning framework. We firstly develop a novel graph augmentation strategy to explore the global structures in graphs. The key idea is to simulate the diffusion processes to sample ordered node sequences named \emph{cascades}, and then use structure inference to recover the augmented graphs from cascades \cite{brugere2018network}. The nodes that co-occurred in the same cascade are assumed to have higher similarities, which will be endowed with higher probabilities of connections in the augmented graphs, and vice versa. The benefits of our augmentation strategy are two-fold: First, the inference process considers intra-graph contrastive information during the optimization, which equips the learned graph structures with more discriminative ability in a global view. Second, compared with previous methods, we build an optimization problem to construct the augmented graphs in a data-driven fashion, where no priors are required. 

With the introduction of the graph augmentation strategy for structural information improvement, we proceed to further leverage the label information to improve graph contrastive learning. To improve the contrastive power, we propose to use a supervised contrastive loss to replace the original self-supervised loss. The label information is injected into the contrastive learning pipeline where the one-vs-one or one-vs-many contrastive paradigms are extended to many-vs-many ones. The introduction of labels also improves the mining of graph patterns that are hard to be distinguished by classifiers \cite{khosla2020supervised}. The supervised contrastive loss with structure inference can be easily integrated into hierarchical GNNs to accomplish graph classifications, where the supervised contrastive loss can be jointly optimized with the general classification loss over augmented graphs. The contributions of this paper can be summarized as follows,
\begin{enumerate}
    \item We propose a novel graph classification method that integrates supervised contrastive learning with structure inference, which improves the contrastive power from the structure and label sides.
    \item We propose a data-driven graph augmentation strategy based on structure inference, where the intra-graph contrastive information is explored to facilitate the graph classification.
    \item We extend the self-supervised graph contrastive learning to a supervised setting where the label information is introduced to further improve the contrastive power.
    \item Experiments on graph classification over several benchmark datasets show the effectiveness of the proposed model.
\end{enumerate}

\section{Related Works}

\subsection{Graph Neural Networks}
Graph neural networks are firstly proposed as a form of recursive networks \cite{scarselli2008graph}. Later methods mainly follow a message passing scheme where the node features can be updated by iteratively aggregating features from neighbor nodes \cite{gilmer2017neural}. Many representative methods can be formulated within the framework, including vanilla GCN \cite{GCNmodel}, GraphSAGE \cite{hamilton2017inductive}, and GAT \cite{velivckovic2018graph}. Limited by the over-smoothing issue, graph neural networks usually suffer from performance degradation when increasing the number of neural layers. To improve the feature extraction abilities of GNNs, later methods focus on extending the local receptive field to encode high-order structures. For example, APPNP \cite{klicpera2019predict} utilize Personalized Page Rank to obtain a better locality for a target node.

Another line of methods focuses on global graph representation learning, which can be directly used for graph classification tasks \cite{zhang2018end, zhu2020gssnn}. The key notion is developing a multi-scale feature extraction pipeline to leverage different levels of topological information in graphs \cite{li2019semi, xuan2019subgraph, yang2021selfsagcn, dai2021hyperbolic}. GIN is proposed as a graph classification method that is as powerful as the Weisfeiler-Lehman Isomorphism test \cite{xu2018powerful}. To accomplish graph-level classifications, the readout functions that produce graph embeddings should be translation-invariant and insensible to matrix conversions. Graph pooling is a commonly-used method in GNNs to generate the translation-invariant embedding for a whole graph. A pooling method usually generates coarser graphs with new node sets by summarizing the node features and topology information \cite{ying2018hierarchical}. A simple strategy is to select a subset of nodes from the original graph where the selected nodes are of larger weights according to certain criteria \cite{zhang2019hierarchical, li2020graph}. For example, the attention mechanism is usually used to determine the subset of nodes \cite{li2019semi, baek2021accurate, lee2019self}.
 
\subsection{Contrastive Learning}
Self-supervised learning is a promising paradigm that designs pre-text tasks to learn generalizable features, where the labels are constructed from the data itself \cite{liu2021self}. Contrastive learning is a branch of self-supervised learning, which learns to compare the similarities of samples. Maximizing mutual information has been widely used in contrastive learning. Deep InfoMax \cite{hjelm2018learning} attempts to maximize the mutual information between the local patches and their global contexts in images. The main assumption is that the global features of images are supposed to have high similarities with their own local patch features, but have low similarities with local patches from other images. Data augmentations are essential in contrastive learning which can generate different views from data. AutoAugment \cite{cubuk2019autoaugment} automatically discovers data augmentation methods to improve the classification performance.

Contrastive learning has also been successfully applied to graph neural networks. Similar to images, it is natural to build local and global contrastive learning objectives. DGI \cite{velickovic2019deep} firstly introduces the mutual information maximization into graph neural networks, which maximizes the mutual information between nodes and the whole graph. By utilizing different scales of structures in graphs, contrastive learning can equip GNNs with different properties. On the sub-graph level, GCC \cite{qiu2020gcc} introduces the InfoCE loss to implement contrastive learning. SAGE \cite{li2019semi} separately leverages graph-instance level and hierarchical graph level information to accomplish the graph classification task. Other scales for contrastive learning include graph-graph level \cite{you2020graph}, graph-patch level \cite{sun2020infograph}, graph-node level \cite{hassani2020contrastive}, and node-node level \cite{di2020mutual}. GXN \cite{li2020graph} develops a cross-scale layer to enable feature exchanges. It can improve the performance when serving as a graph pooling method.

\section{Our Method}
We begin our discussion by giving the notations used in this paper. A graph can be represented as $ G = (V,E) $, where $V$ is the node set, and $E$ is the edge set. The adjacent matrix and the attribute matrix of the graph are denoted as $\mathbf{A} \in \mathbb{R}^ {n\times n}$ and $ \mathbf{X} \in \mathbb{R}^{n \times d}$, where $n$ is the number of nodes, and $d$ is the dimension of node attributes. The neighbor set for node $v_{i}$ is denoted as $\mathcal{N}(v_{i})$. Given a dataset containing $m$ graphs with labels $\mathcal{D}=\{(G_{i},\mathbf{y}_{i})\}_{i=1}^{m}$, the goal of graph classification is to learn a prediction function $F$ from the graph space $\mathcal{G}$ to the label space $\mathcal{Y}$.

\begin{figure*}[htbp]
  \centering
  \includegraphics[width=1.0\linewidth]{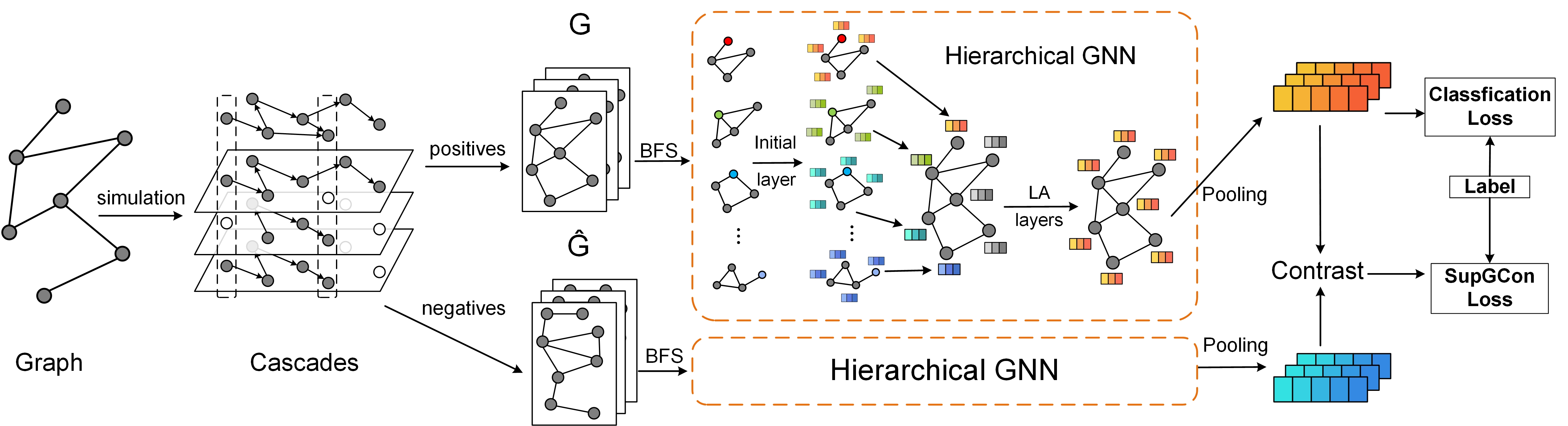}
   \caption{An illustration of the proposed method, SupCosine. First, for each graph in the dataset, we simulate the diffusion processes to generate cascades that capture node similarities. The positive graph $G$ is recovered from the cascades by structure inference with augmentations. The negative graph $\hat{G}$ is sampled from the recovered graphs. Second, after obtaining the batches of graphs, the positive batch and the negative batch are passed through the hierarchical graph neural networks to produce graph-level embeddings. Here, the Initial layer represents the initialization layer that generates initial node features with sub-graph patterns, LA layers denote multiple local aggregation layers to further update node features, BFS denotes the Breadth-First-Search algorithm to generate sub-graphs. Finally, embeddings of positives in the batch are utilized for supervised graph classification loss. The embeddings of positives and negatives are served as contrastive pairs for the supervised graph contrastive loss.}
   \label{framework}
\end{figure*}

Our method is proposed to improve the contrastive power of graph neural networks for graph classification. The schematic illustration of the method is presented in Figure \ref{framework}. The details of hierarchical graph neural networks used for graph feature learning are firstly presented. Then we introduce the two modules in our model: the supervised graph contrastive learning, and the structure inference for graph augmentation. 

\subsection{Hierarchical Graph Neural Networks}

The skeleton of the proposed model is realized by a hierarchical graph neural network. In general, the process to obtain graph embeddings can be broken down into local aggregation functions to update node features and graph pooling functions to hierarchically summarize node features.  



\textbf{Sub-graph Feature Initialization.} Since the local aggregation functions can only accumulate features from direct neighbors, we set the first layer of our method as a sub-graph feature initialization layer to aggregate sub-graph level structures. This layer serves as a preparation for the later local aggregation layers. 

The core of sub-graph feature initialization is to build a reasonable sub-graph for each node. We resort to Breath-First-Search (BFS) following \cite{sun2021sugar} to construct sub-graphs. For each node $v_{i}$, we can obtain its corresponding sub-graph $g_{i}$ with node set $\mathcal{S}_{\text{BFS}}(v_{i})$. To facilitate computation, the size of $\mathcal{S}_{\text{BFS}}(\cdot)$ is controlled by a hyper-parameter $\beta$ which denotes the maximal nodes BFS can generate. Based on the sub-graphs, we can initialize the node features by local aggregation functions:
\begin{align}
\mathbf{a}^{(0)}_{v} = &\operatorname{AGGREGATE}(\mathbf{x}_{u} \mid u \in \mathcal{S}_{\text{BFS}}(v)),\label{initial_agg} \\
\mathbf{h}^{(0)}_{v} = &\operatorname{COMBINE}(\mathbf{x}_{v}, \mathbf{a}^{(0)}_{v}), \label{initial_comb}
\end{align}
where $\mathbf{h}^{(0)}$ is the initialized node features for node $v_{i}$, $\operatorname{AGGREGATE}$ is the aggregation function that accumulates neighbor features, and $\operatorname{COMBINE}$ is the combine function that combines the neighbor features and the features from the node itself. We can then obtain the output of the first layer $\mathbf{H}^{(0)}$. Notice that the number of nodes generated by BFS is relatively small, which ensures that the computation complexity remains low compared with a complete aggregation layer.

\textbf{Multi-layer Local Aggregations.} To improve the feature extraction ability of the proposed model, we further stack several local aggregation layers after the initialization layer. The layer-wise updating rule can be denoted as:
\begin{equation}
\small
\mathbf{h}^{(l)}_{v} = \operatorname{MLP}^{(l)}\left( (1+\gamma^{(l)})\mathbf{h}_{v}^{(l-1)} +\sum_{u \in \mathcal{N}(v_{i})}\mathbf{h}_{u}^{(l-1)}\right),
\label{GNN}
\end{equation}
where $l \geq 1$ is the layer index, $\gamma$ is a small number, $\mathcal{N}(v_{i})$ is the neighbor set of node $v_{i}$, and $\operatorname{MLP}^{(l)}$ is the multi-layer perception used for feature transformation. After $L$ local aggregation layers, the node representations are updated as $\mathbf{Z} = \mathbf{H}^{(L)} \in \mathbb{R}^{n \times d_{n}}$ where $d_{n}$ is the node feature dimension in the $L$-th layer. 

\textbf{Graph Feature Learning.}  After obtaining the node features, the next goal is to summarize the node features to formulate the graph-level representations. To extract hierarchical graph structures, we first apply a graph pooling layer to retrench a original graphs $G$ as a coarser graph $\widetilde{G}$. Concretely, consider the node set $V=\{v_{i}\}_{i=1}^{n}$, we use a global ranking function with a linear projection operation to obtain the coarser graph,
\begin{equation}
\label{pooltopk}
    idx = \operatorname{TOP}_{k} (\frac{\mathbf{Z}\mathbf{w}}{\|\mathbf{w}\|}),
\end{equation}
where $idx$ denotes the indices of nodes for the coarsened graph, $\mathbf{w} \in \mathbb{R}^{d_{n}\times 1}$ is a vector of learnable parameters that maps node features into 1 dimension, and $\operatorname{TOP}_{k}(\cdot)$ is the top-$k$ function that selects $k$ largest values from the projected results. The results of the linear projection are normalized by the norm of the parameters.

According to $idx$, we can construct the coarser graph $\widetilde{G}$ with node set $\widetilde{V}$ and edge set $\widetilde{E}$. To further leverage the sub-graph information, we use the sub-graphs generated by BFS to build the coarser graph, where each selected sub-graph is treated as a super node. The edges of the coarser graph are determined by the number of overlapping nodes between two sub-graphs.
\begin{align}
    \widetilde{V} = & \{ \widetilde{v}_{i}; \widetilde{v}_{i}=g_{i},  \forall i \in idx \},
    \label{corase_V}\\
    \widetilde{E} = & \{\widetilde{e}_{i,j}; \forall  |\mathcal{S}_{BFS}(v_{i}) \cap \mathcal{S}_{BFS}(v_{j})| \geq \varepsilon \},
    \label{corase_E}
\end{align}
where $\widetilde{v}$ and $\widetilde{e}$ denote the node and the edge in the coarser graph, and $\varepsilon$ is the threshold parameter. The representations of the super nodes are calculated by a linear transformation,
\begin{align}
    \widetilde{\mathbf{z}}_{i} = \sum_{\widetilde{e}_{i,j} \in \widetilde{E}} \mathbf{W}_{g}\mathbf{z}_{i},
    \label{corase_X}
\end{align}
where $\widetilde{\mathbf{z}}_{i}$ is the feature vector of the super node $\widetilde{v}_{i}$, $\mathbf{W}_{g} \in \mathbb{R}^{d_{s} \times d_{n}}$ is the parameter matrix, and $d_{s}$ is feature dimension for the super nodes.

We can then calculate the graph-level embeddings with respect to the representations of the coarser graph via a readout function,
\begin{align}
\label{embed of graph}
    \mathbf{r} = \operatorname{READOUT}(\{\widetilde{\mathbf{z}}_{i}\}_{i=1}^{k}),
\end{align}
where $\mathbf{r} \in \mathbb{R}^{d_{g} \times 1}$ is the embedding of the graph.

\textbf{Graph Classification.} The optimization objective for the graph classification is a cross-entropy loss which can be written as,
\begin{equation}
\label{gc_loss}
    \mathcal{L}_{gc} = \operatorname{CrossEntropy}(\mathbf{Y}, \mathbf{P}),\\
\end{equation}
where $\mathbf{Y}$ denotes the ground truth labels for graphs, and $\mathbf{p} \in \mathbf{P}$ denotes the predictions obtained from graph embeddings $\mathbf{r}$ via classification layers.

\subsection{Supervised Graph Contrastive Learning}
In this subsection, we introduce the supervised graph contrastive loss that can be used to improve the contrastive power of the proposed model. The essential part of graph contrastive learning is to build contrastive pairs from the graphs. Previous methods sample from different scales of the topological structures to formulate contrastive pairs. Based on the hierarchical graph neural networks, we build our contrastive loss from the global graph embeddings. 

As indicated in Figure \ref{framework}, the positive and negative samples are separately drawn from the augmented graph $G$, and are then fed into the hierarchical graph neural network to obtain graph embeddings. We first introduce the self-supervised contrastive loss based on graph embeddings,
\begin{equation}
\mathcal{L}_{self} = -\sum_{i\in  \Omega}{\operatorname{log}\frac{\operatorname{exp}(\mathbf{r}_{i}\cdot \mathbf{r}_{j(i)}/\tau)}{\sum_{k \in \Gamma(i)}{\operatorname{exp}(\mathbf{r}_{i}\cdot \mathbf{r}_{k}/\tau})}},
\label{SelfCon_loss}
\end{equation}
where $\Omega =\{1,\dots,2m\}$ represents the index set for all positive and negative samples in the data, $\Gamma(i)=\Omega \backslash {i}$, $\tau$ is a positive temperature parameter, $\cdot$ is the dot product, $i$ is the anchor, $j(i)$ denotes the positive sample, and others are negative samples.

Since the label information is available in supervised learning scenarios, more positive samples can be included to improve the contrastive power. Under this scenario, the one-vs-many contrastive learning can be extended to many-vs-many settings. In other words, compared with self-supervised contrastive learning, supervised contrastive learning has more positive and negative data for contrastive learning. To leverage label information, we extend the loss in \cref{SelfCon_loss} to a supervised graph contrastive (SupGCon) loss with the following form \cite{khosla2020supervised}:
\begin{equation}
\small
\mathcal{L}_{sc} = \sum_{i\in \Omega}{\frac{-1}{|\Phi(i)|}\sum_{p\in \Phi(i)}{\operatorname{log}\frac{\operatorname{exp}(\mathbf{r}_{i}\cdot \mathbf{r}_{p}/\tau)}{\sum_{ k \in \Gamma(i)}{\operatorname{exp}(\mathbf{r}_{i}\cdot \mathbf{r}_{k}/\tau})}}},
\label{SupCon_loss}
\end{equation}
where $\Phi(i)$ is the index set for all positive samples in a batch, $|\Phi(i)|$ denotes the number of elements in the set.

\subsection{Structure Inference}
In this subsection, we mainly discuss the graph augmentation strategy based on structure inference. The goal of structure inference is to enhance the existing edges by discovering possible connections between nodes with high similarities. To capture node similarities, we resort to diffusion samplings \cite{shi2019diffusion} that simulate diffusion processes in graphs. It starts from a root node and then activates neighbor nodes at a certain rate. This process can generate ordered node sequences, i.e., \emph{cascades}, that capture node similarities. A cascade $\mathbf{c}^{i}$ is represented as an $n$-dimensional vector $(t_{1}^{i},\cdots,t_{n}^{i})$ where each element represents the time-stamp of the activated time for a given node. Only the time-stamps within a fixed time window $T$ are considered. The time-stamps for inactivated nodes are denoted as $\infty$. A set of cascades is represented as $\mathbf{C}:(\mathbf{c}^{1},\cdots,\mathbf{c}^{q}) \in \mathbb{R}^{q \times n}$ where $q$ is the number of cascades.

To recover the possible connections, we formulate a learning objective that embeds the node similarities in cascades into connection weights \cite{gomez2011uncovering}. Concretely, we use a bunch of observed cascades to infer a weighted connection matrix $\mathbf{M}$ by maximizing the likelihood $\phi(\mathbf{C};\mathbf{M})$. We begin by considering the inference problem of a single cascade $\mathbf{c} \in \mathbf{C}$. Given the time window $[0,T]$, the inference problem for $\mathbf{c}$ is $\phi(\mathbf{c};\mathbf{M})$, which is a joint likelihood of activated nodes (i.e. $t_{i}\leq T$) and inactivated nodes (i.e. $t_{i} > T$). The pair-wise transmission from $v_{i}$ to $v_{j}$ can be modeled by a network model, e.g., exponential model,
\begin{equation}
\small
f(t_{j}|t_{i};\mathbf{M}_{i,j})=\left\{
\begin{array}{lcl}
\mathbf{M}_{i,j} \cdot e^{-\mathbf{M}_{i,j}(t_{j}-t_{i})}, &  {if \ t_{i}<t_{j}}, \\
0,                                       &  {otherwise}.
\end{array}
\right.
\end{equation}
Then, the probability that $v_{j}$ is activated by $v_{i}$ can be represented as the joint probability that $v_{j}$ is activated by $v_{i}$ and not being activated by other already activated nodes in the cascade. $\phi(t_{j}|t_{i};\mathbf{M})$ can be denoted as,
\begin{equation}\label{pair_likeli}
\small
\phi(t_{j}|t_{i};\mathbf{M}) =  
   f(t_{j}|t_{i};\mathbf{M}_{i,j})
\prod_{\begin{subarray}{c} t_{k} \neq t_{i} \\ t_{k}< t_{j}\end{subarray}}
\Big{(} 1- f(t_{j}|t_{k};\mathbf{M}) \Big{)},
\end{equation}
From Equation \eqref{pair_likeli}, we can calculate $\phi(t_{j};\mathbf{M})$ by summing up the likelihood denoting $v_{i}$ ($t_{i}<t_{j}$) is the first node that activates $v_{j}$, 
\begin{equation}
\footnotesize
\begin{split}
\phi(t_{j};\mathbf{M}) = &
\sum_{\begin{subarray}{l}
      t_{i} < t_{j}
      \end{subarray}}
      \phi(t_{j}|t_{i};\mathbf{M})  \\
=&
\sum_{\begin{subarray}{l}
      t_{i} < t_{j}
      \end{subarray}}
\Big{[} f(t_{j}|t_{i};\mathbf{M})
\prod_{\begin{subarray}{l}
      t_{k} \neq t_{i} \\ t_{k} < t_{j}
      \end{subarray}}
\Big{(} 1-f(t_{j}|t_{k},\mathbf{M}) \Big{)} \Big{]}. 
\end{split}
\end{equation}
Then, the likelihood of observing all activations in a cascade $\phi(\mathbf{c}^{\leq T};\mathbf{M})$ is the joint probability of $\phi(t_{j};\mathbf{M})$ for all $t_{j} \leq T$,

\begin{equation}
\footnotesize
\begin{split}
\phi(\mathbf{c}^{\leq T};\mathbf{M})
= \prod_{t_{j} \leq T}
\Bigg{(}
\sum_{\begin{subarray}{l} t_{i} < t_{j} \end{subarray}}
      \frac{f(t_{j}|t_{i};\mathbf{M})}{1- f(t_{j}|t_{i};\mathbf{M})}
\prod_{\begin{subarray}{l} t_{k} < t_{j} \end{subarray}}
        \Big{(} 1-f(t_{j}|t_{k};\mathbf{M}) \Big{)}
\Bigg{)}. 
\end{split}
\end{equation}

The inactivated nodes where $t_{l}>T$ provide contrastive information when inferring the possible edges. Consequently, the likelihood function for a cascade $\mathbf{c}$ can be represented as,

\begin{equation}
\footnotesize
\begin{split}
\phi(\mathbf{c};\mathbf{M})
= & \phi(c^{\leq T};\mathbf{M}) \times  \phi(c^{> T};\mathbf{M})  \\
= & \prod_{t_{j} \leq T}
\Bigg{(}
\sum_{\begin{subarray}{l} t_{i} < t_{j} \end{subarray}}
\frac{f(t_{j}|t_{i};\mathbf{M})}{ 1-f(t_{j}|t_{i};\mathbf{M})}
\prod_{\begin{subarray}{l} t_{k} < t_{j} \end{subarray}}
       \Big{(} 1-f(t_{j}|t_{k};\mathbf{M}) \Big{)}
\Bigg{)}
\\
&
\times
\Bigg{(} \prod_{\begin{subarray}{l} t_{l} > T \end{subarray}}
         \prod_{\begin{subarray}{l} t_{j} \leq T \end{subarray}}
         \Big{(}1-f(t_{l}|t_{j};\mathbf{M}) \Big{)}
\Bigg{)},
\end{split}
\end{equation}

where $\phi(\mathbf{c}^{>T};\mathbf{M})$ denotes nodes $v_{l}$ is not activated by already activated nodes $v_{j}$ in the cascade.

Considering the independence observation of all cascades, the estimation over all observed cascades turns to obtain $f(\mathbf{C};\mathbf{M}) = \prod_{\mathbf{c} \in \mathbf{C}} f(\mathbf{c};\mathbf{M})$. The objective that required to be optimized is denoted as,
\begin{equation}\label{netinf_loss}
\mathop{\max}\limits_{\mathbf{M} \geq 0} \sum_{\mathbf{c} \in \mathbf{C}} \log f(\mathbf{c};\mathbf{M}).
\end{equation}
Notice that the resulted connection matrix $\mathbf{M}$ may contain noisy connections. To this end, we use a parameter $\xi$ to denote the number of edges added to the original graph. we can formulate a new adjacent matrix by combining the original adjacent matrix and the learned connection matrix via $\mathbf{A}' = \mathbf{A} \oplus \phi_{\xi}( \mathbf{M})$, where $\oplus$ is the element-wise addition, and $\phi_{\xi}$ is the function to select augmented edges.

\begin{algorithm}
\caption{The Training Process of SupCosine}
\label{supercosine}
\KwIn{Graphs with labels $\mathcal{D}=\{(G_{i},\mathbf{y}_{i})\}_{i=1}^{m}$; coefficient of SupGCon loss $\lambda$; Up-limit of BFS $\beta$; Number of epochs $epochs$, batch size $B$}
\KwOut{Graph labels $\mathcal{Y}$, Model parameters $\mathbf{W}$}
\For{$G \in \mathcal{D}$}{
Cascade Set $\mathbf{C} \gets\operatorname{ simulation}(\mathbf{A})$\;
$\mathbf{M} \gets $\cref{netinf_loss}; \hfill \tcp{\footnotesize{Structure Inference}}
$\mathbf{A}' = \mathbf{A} \oplus \phi_{\xi}( \mathbf{M})$ \;
}
$\mathcal{B},\mathcal{\hat{B}} \gets \operatorname{sampling}(\mathcal{D})  $\;
\For{epoch $<$ epochs}{
\For{$G,\hat{G} \in \mathcal{B},\mathcal{\hat{B}}$}{
$\mathbf{H}^{0},\hat{\mathbf{H}^{0}} \gets $\cref{initial_comb};\hfill\tcp{\footnotesize{Initialization}}
$\mathbf{Z},\hat{\mathbf{Z}} \gets$ \cref{GNN}\;
$idx \gets$ \cref{pooltopk}\;
$\widetilde{\mathbf{Z}} \gets$ \cref{corase_V}, \cref{corase_E}, and \cref{corase_X}\;
$\mathbf{r}\gets$\cref{embed of graph};\hfill \tcp{\footnotesize{Top k nodes}}
}
$\mathcal{L}_{gc} \gets$ \cref{gc_loss};\hfill \tcp{\footnotesize{Cross Entropy Loss}}
$\mathcal{L}_{sc} \gets$ \cref{SupCon_loss};\hfill \tcp{\footnotesize{SupGCon Loss}}
$\mathcal{L}\gets$ \cref{total_loss}\;
Update $\mathbf{W} \gets$ Backward($\mathcal{L}$)\;
}
\end{algorithm}

\subsection{Algorithm}
Our overall optimization objective is the combination of the graph classification loss and the supervised graph contrastive loss,
\begin{equation}
\label{total_loss}
    \mathcal{L} = \mathcal{L}_{gc} + \lambda \mathcal{L}_{sc},
\end{equation}
where $\lambda$ is a positive hyper-parameter. By jointly optimizing the graph classification loss and the supervised graph contrastive loss, the learned graph embeddings are more discriminative for graph classification. The overall algorithm of the proposed model is listed in Algorithm \ref{supercosine}.

\section{Experiments}
To evaluate the effectiveness of the supervised graph contrastive learning with structure inference in graph neural networks, we conduct experiments to show the improvements compared with previous graph classification methods. 
 
\subsection{Data Description}
Four benchmark datasets, MUTAG, PTC, IMDB-BINARY, and PROTEINS, are used for evaluation. MUTAG and PTC are molecule graph datasets, IMDB-BINARY is a social network dataset, and PROTEINS is a bioinformatics graph dataset. Detailed information of the datasets is summarized in Table \ref{data}. 

\begin{table}[htbp]
\centering
\small
\begin{tabular}{lcccc}
\hline
Datasets & MUTAG & PTC   & IMDBB & PROTEINS \\ 
\hline
$\#$Classes   & 2      & 2      & 2     & 2        \\
$\#$Graphs    & 188    & 344    &1000   & 1113      \\
Avg. Nodes    & 17.93  & 25.56  & 19.77 & 39.06    \\
Max. Nodes    & 28     & 109    &136    & 620       \\
$\#$Node Labels   & 7      & 19     & 1     & 3        \\
\hline
\end{tabular}
\caption{The statistics of the datasets.}
\label{data}
\end{table}

\subsection{Baseline Methods}
For comprehensive comparisons, we compare the proposed method with three types of graph methods: 

(1) Graph kernel methods: Weisfeiler-Lehman Subtree Kernel (WL), Graphlet kernel (GK), Deep Graph Kernels (DGK), graph kernel based on Wasserstein embedding (WEGL), and graph kernel based on returned random walks (RetGK). 

(2) Graph neural network methods: GraphSAGE is developed for node classification, max pooling is used to obtain the graph level representations. GIN, DAGCN, PPGN, CapsGNN and SGN are developed for graph-level representations or classifications. SAGPool, NDP and GMT focus on the graph pooling methods which can learn hierarchical graph structures.

(3) Graph contrastive learning methods: GraphCL and M-GCL are based on graph augmentations, InfoGraph, GXN, sGIN and SUGAR are based on mutual information maximization.

\begin{table}[htbp]
\footnotesize
\setlength\tabcolsep{3pt}
\centering
\begin{tabular}{ccccc}
Methods  & MUTAG       & PTC         & PROTEIN     & IMDBB       \\ 
\hline   
WL \cite{shervashidze2011weisfeiler}  & 83.8\(\pm\)1.5   & -  & 74.7\(\pm\)0.5   & 73.4\(\pm\)4.6           \\
GK \cite{shervashidze2009efficient}   & 83.5\(\pm\)0.6     & 59.7\(\pm\)0.3 & - & - \\
DGK \cite{yanardag2015deep}           & 87.4\(\pm\)2.7 & 60.1\(\pm\)2.6 & 75.7\(\pm\)0.5 & -   \\ 
WEGL \cite{kolouri2021wasserstein}    &- &67.5\(\pm\)7.7 &76.5\(\pm\)4.2 &75.4\(\pm\)5.0 \\
RetGK \cite{zhang2018retgk}    &90.3\(\pm\)1.1 &67.9\(\pm\)1.4 &75.8\(\pm\)0.6 &72.3\(\pm\)0.6 \\
\hline
GraphSAGE\cite{hamilton2017inductive}  & 79.8\(\pm\)13.9 & -       & 65.9\(\pm\)2.7 & 72.4\(\pm\)3.6 \\ 
GIN\cite{xu2018powerful}     & 89.4\(\pm\)5.6    & 64.6\(\pm\)7.0   & 76.2\(\pm\)2.8   & 75.1\(\pm\)5.1   \\
DAGCN\cite{chen2019dagcn}   &87.2\(\pm\)6.1 &62.9\(\pm\)9.6 &76.3\(\pm\)4.3 &- \\
PPGN\cite{vignac2020building}    &90.6\(\pm\)8.7 &66.2\(\pm\)6.5 &77.2\(\pm\)4.7 &73.0\(\pm\)5.8 \\
CapsGNN\cite{xinyi2018capsule}     &86.7\(\pm\)6.9 &- &76.3\(\pm\)3.6 &73.1\(\pm\)4.8 \\
SGN\cite{xuan2019subgraph}     &89.5\(\pm\)7.4 &64.1\(\pm\)3.7 &76.3\(\pm\)4.1 &76.5\(\pm\)5.7 \\
SAGPool\cite{lee2019self}    &76.8\(\pm\)2.1 &- &72.0\(\pm\)1.1 &72.2\(\pm\)0.9 \\
NDP\cite{bianchi2020hierarchical}     &87.9\(\pm\)5.7 &- &73.4\(\pm\)0.8 &-  \\
GMT\cite{baek2021accurate}    &83.4\(\pm\)1.3 &- &75.1\(\pm\)0.6 &73.48\(\pm\)0.8 \\
\hline
GraphCL \cite{you2020graph} &86.8\(\pm\)1.3 &- &74.4\(\pm\)0.5 &71.1\(\pm\)0.4 \\
InfoGraph \cite{sun2020infograph} & 89.0\(\pm\)1.1 & 61.7\(\pm\)1.4 & -  & 73.0\(\pm\)0.9 \\
M-GCL \cite{hassani2020contrastive} &89.7\(\pm\)1.1 &62.5\(\pm\)1.7 &- &74.2\(\pm\)0.7\\
GXN \cite{li2020graph}   &86.1\(\pm\)8.3 &63.5\(\pm\)5.8 &79.9\(\pm\)4.1 &78.6\(\pm\)2.3\\
sGIN\cite{di2020mutual}    & 94.1\(\pm\)2.7 & 73.6\(\pm\)4.3 & 79.0\(\pm\)3.2 & 77.9\(\pm\)4.3 \\
SUGAR \cite{sun2021sugar}   & 96.7\(\pm\)4.6 & 77.5\(\pm\)2.8 & \textbf{81.3\(\pm\)0.9} & 73.0\(\pm\)3.5  \\ 
\hline
SupCosine & \textbf{98.3\(\pm\)2.5} & \textbf{87.8\(\pm\)10.4} & \textbf{80.0\(\pm\)3.6} & \textbf{83.0\(\pm\)3.2} \\
\hline
\end{tabular}
\caption{Classification results compared with baseline methods.}
\label{comparision}
\end{table}

\subsection{Implementation Details}
We use the RMSprop optimizer with $\alpha=0.9$, and weight decay $0.001$. We adopt 5 local aggregation layers with 2 MLP layers. The dimensions of all hidden layers are set as 16. The aggregation type is the sum function. The temperature $\tau$ of supervised graph contrastive loss is set as 0.07.  In structure inference, the time window $T$ is set as 10, and the type of diffusion is set as the exponential model. The coefficient $\lambda$ is selected from [0.005, 0.01, 0.02, 0.03]. For each dataset, the upper limit of BFS $\beta$ is selected from \([3,5,7]\). For larger datasets, the number of augmented edges is selected from [3, 5, 7], and from [1, 2, 3] for other datasets.

\subsection{Classification Results}
In this subsection, we compare the classification accuracy of our method and several strong baseline methods under four datasets, and report the results in \cref{comparision}. The best results are marked in bold, and '-' denotes that the results are not available. 

As shown in \cref{comparision}, the accuracy of our method shows strong competitiveness with stable standard variance, and achieves SOTA at most datasets. In detail, our method achieves 1.0\%, 10\% and 4.4\% improvements compared with the second-best methods in MUTAG, PTC, and IMDBBINARY respectively. The performance on PROTEIN also achieves SOTA. Specially, we focus on our performance improvements compared with recent graph contrastive learning methods. The comparison shows the effectiveness of our strategies. When comparing with the most recent method, SUGAR, our method can obtain a maximal 10\% improvement in several datasets.

\begin{figure}[htbp]
  \centering
 \includegraphics[width=0.9\linewidth]{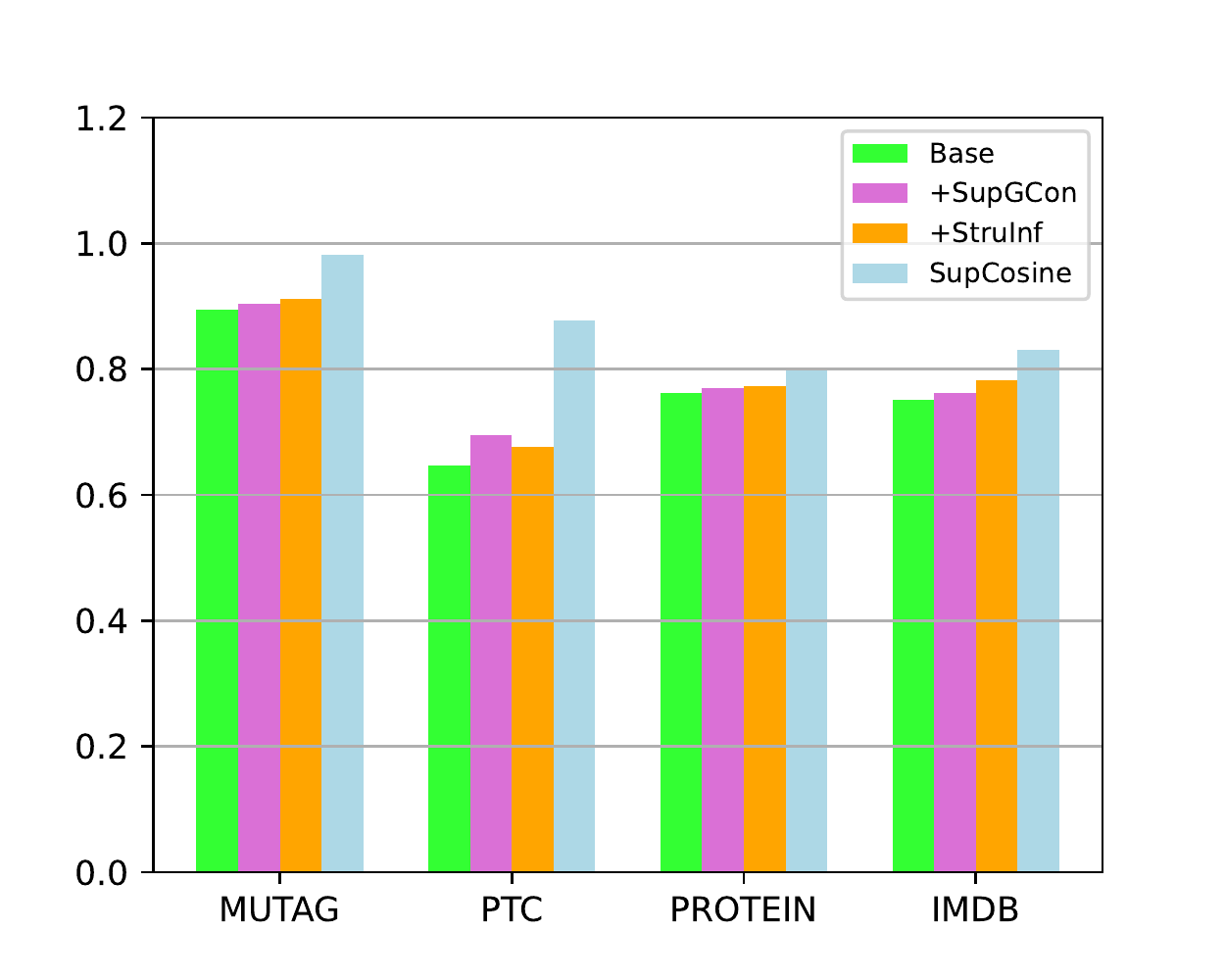}
   \caption{Ablation results.}
   \label{ablation}
\end{figure}

\subsection{Ablation Studies}

In this subsection, we conduct ablation experiments to separately demonstrate the roles of structure inference (StruInf) and the supervised graph contrastive loss (SupGCon). The classification results on four datasets, MUTAG, PTC, PROTEIN and IMDM-BINARY, are shown in Figure \ref{ablation}. \emph{Base} is the method that removes SupGCon and StruInf from SupCosine. \emph{+StruInf} and \emph{+SupGCon} denote the models that add the corresponding components in the \emph{Base} model.

On the one hand, the results show that both strategies can separately improve the classification performances compared with the Base model, which shows the effectiveness of these modules. In most cases, the performance gains obtained by StruInf are slightly higher than SupGCon. On the other hand, SupGCon and StruInf have nicely incorporated within graph neural networks, so that the classification performance has been largely improved when both strategies are employed. The superiority of performance when combining SupGCon and StruInf is more significant in PTC dataset.

\begin{figure}[htbp]
  \centering
 \includegraphics[width=0.9\linewidth]{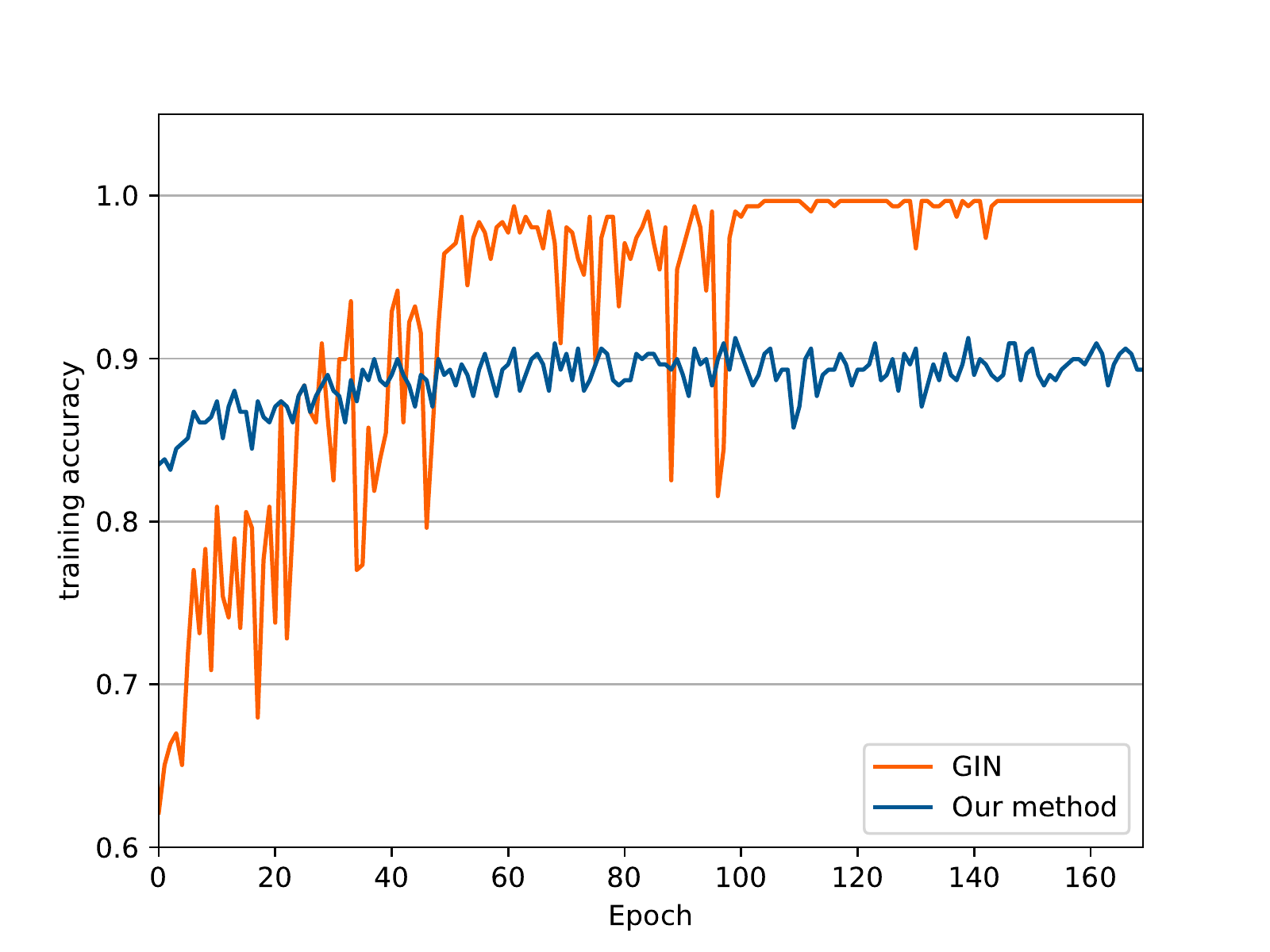}
   \caption{The training curves of accuracy of SupCosine on PTC dataset.}
   \label{acc}
\end{figure}

\begin{figure}[htbp]
\centering
 \includegraphics[width=0.9\linewidth]{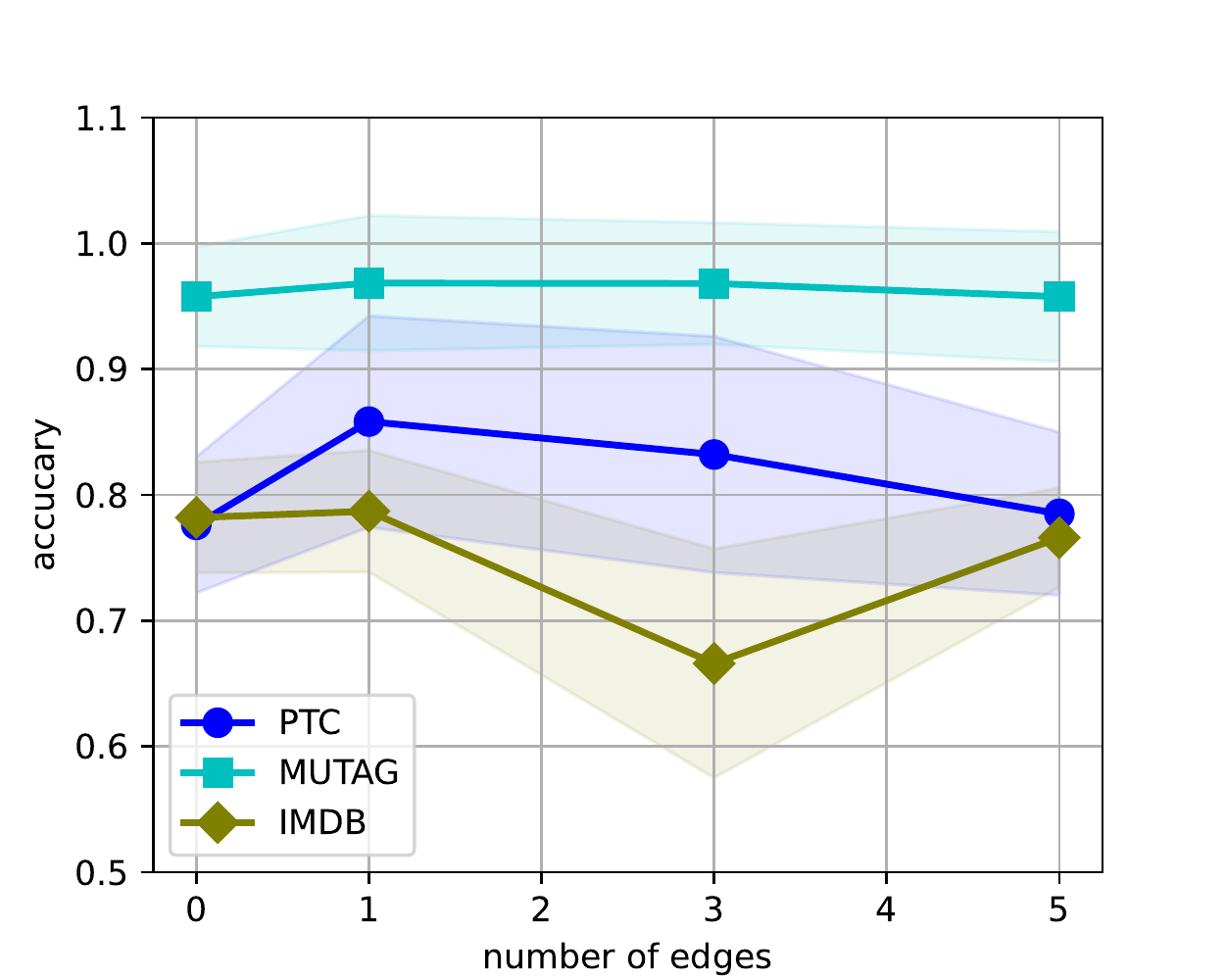}
   \caption{The performance of different numbers of edges added during the structure inference.}
   \label{topk}
\end{figure}

\subsection{Model Analysis}
In this subsection, we give a more detailed analysis of our model. First, we plot the training accuracy of our method and GIN \cite{xu2018powerful} in Figure \ref{acc}. As shown in this figure, our method starts from high accuracies and is more stable during the training process. In contrast, the training curve of GIN fluctuates dynamically with the training proceeds. The training curve of GIN reaches nearly 100\% after 100 epochs, which may also suffer from over-fitting risks. The steady training process of our method mainly benefits from the structure inference process which refines the salient connections within the graphs. Besides, the combination of supervised graph contrastive learning equips the graph embeddings more discriminative ability, which also facilitates the convergences of the training process. 

\begin{figure}[htbp]
\centering
 \includegraphics[width=0.9\linewidth]{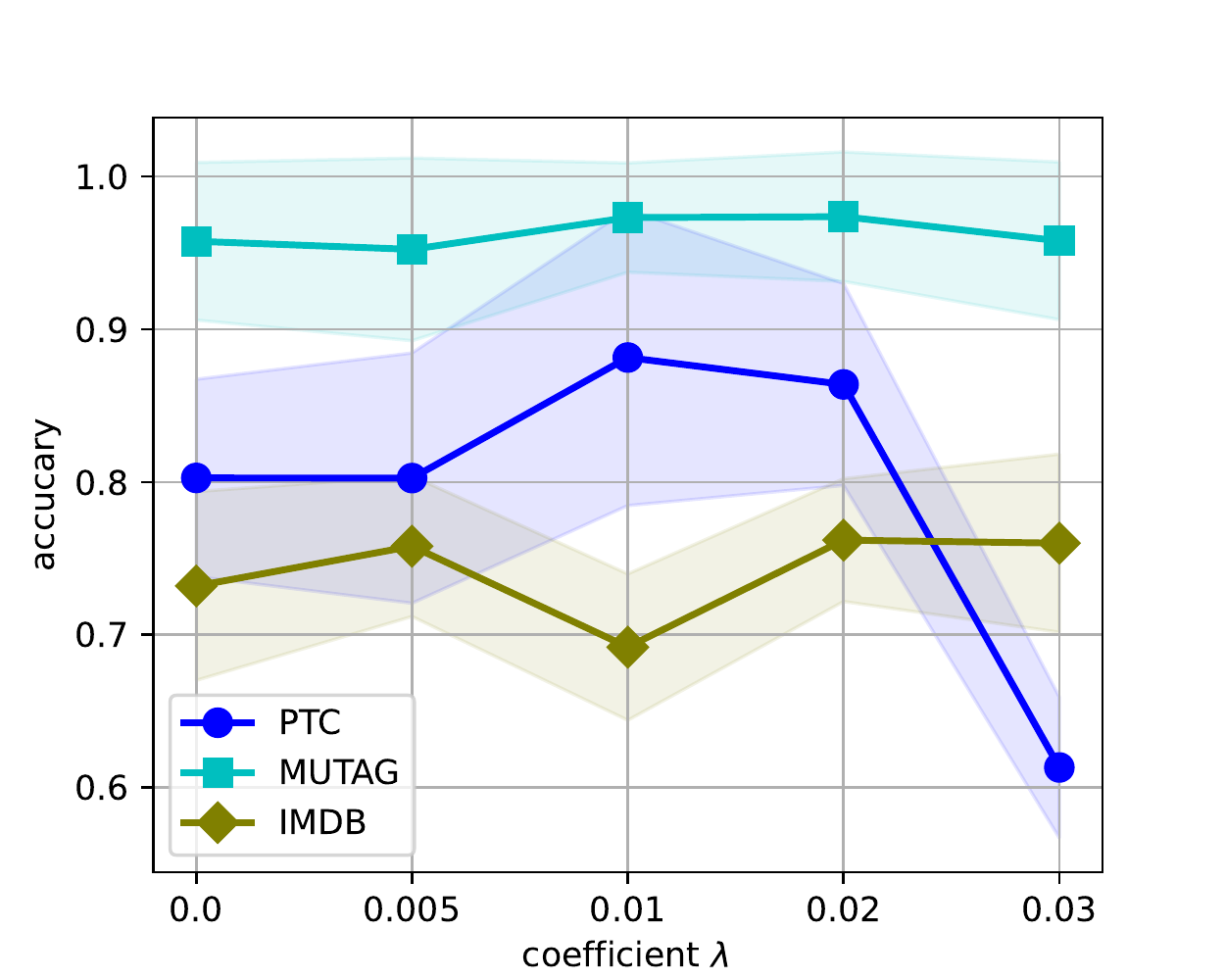}
   \caption{The performance of different $\lambda$ for SupGCon loss.}
   \label{MI loss}
\end{figure}

\begin{figure}[htbp]
\centering
 \includegraphics[width=0.9\linewidth]{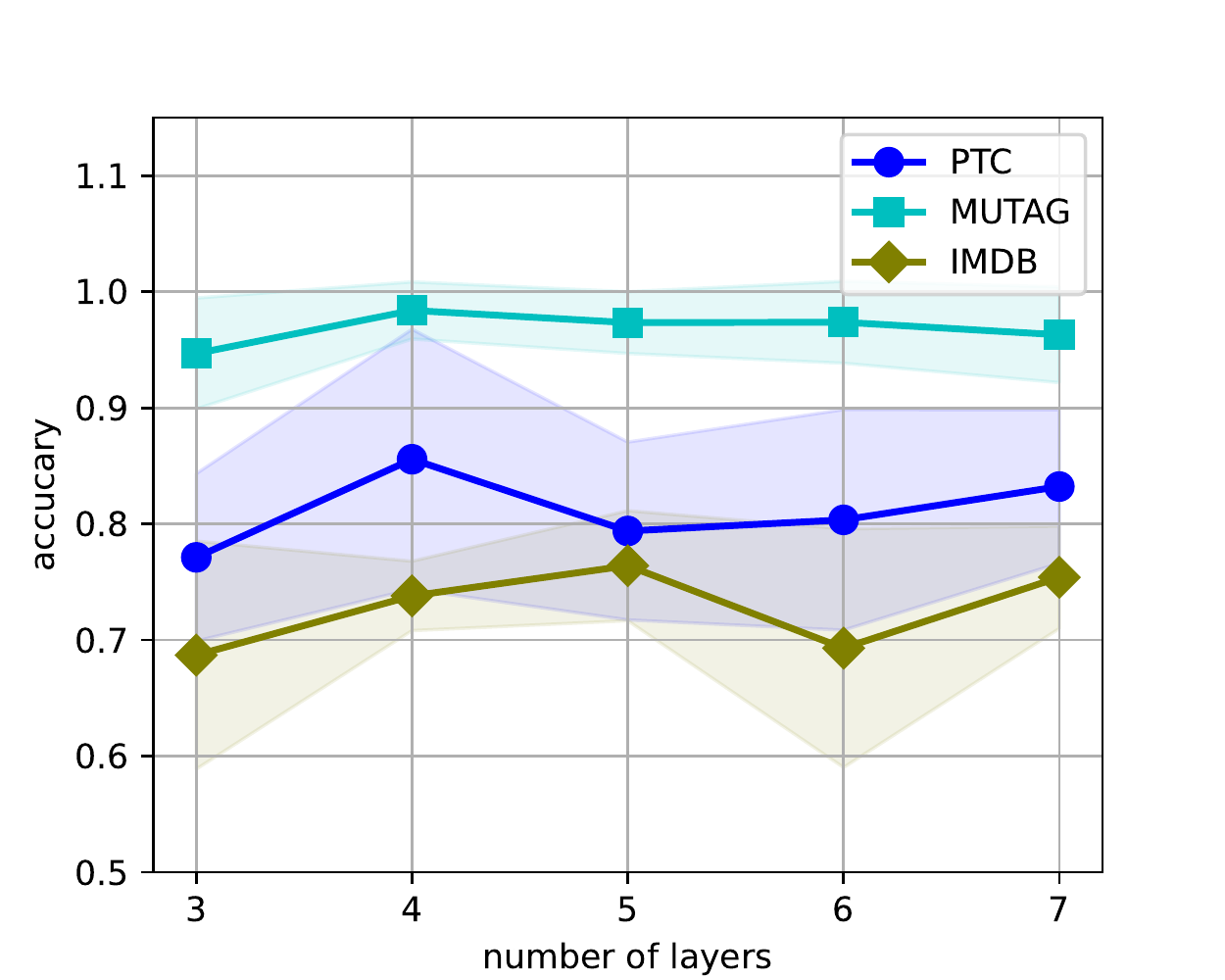}
   \caption{The performance of different local aggregation layers.}
   \label{layers}
\end{figure}

Secondly, we analyze the influences of three representative hyper-parameters used in our model under MUTAG, PTC and IMDB-BINARY datasets. Figure \ref{topk} plots the number of nodes that are added in the structure inference phase. For all three datasets, the highest accuracy is achieved when the number is set as 1. Since most graphs in the datasets are relatively small, when too many edges are added, the graph features are prone to be smoothed, which are less discriminative when performing the classifications. Figure \ref{MI loss} shows the influences of coefficient $\lambda$ that controls the weight of SupGCon loss during the optimization. The results show that the optimal $\lambda$ differs with respect to different datasets. For example, the optimal value is 0.01 for PTC, and is 0.02 for IMDB-BINARY. Figure \ref{layers} gives the performance of our method with different numbers of local aggregation layers. For MUTAG and PTC, 4 layers obtain the best performance. For IMDB-BINARY, 5 layers achieve the best results. With a few layers, the model cannot fully extract the structural information. While too many layers may lead to over-smoothing with indistinguishable node embeddings that ultimately degrade the classification performance.

\section{Conclusions}
In this paper, we studied the challenges of graph contrastive learning in graph classification, and pointed out that the structure information and label information have not been fully explored in existing methods. From this perspective, we proposed a supervised contrastive learning model with structure inference to incorporate the above information within a hierarchical graph neural network framework. Compared with previous graph contrastive learning methods, our advantages are two-fold: First, we introduced a data-driven graph augmentation strategy to enhance the existing edges, from which the contrastive pairs can carry more topological information. Second, we extended the self-supervised contrastive loss to a supervised setting by integrating the label information within the loss. The one-vs-many contrast was then extended to many-vs-many contrasts so that the contrastive power of the model can be largely improved. 

{\small
\bibliographystyle{ieee_fullname}
\bibliography{citation}

\begin{thebibliography}{10}\itemsep=-1pt

\bibitem{baek2021accurate}
Jinheon Baek, Minki Kang, and Sung~Ju Hwang.
\newblock Accurate learning of graph representations with graph multiset
  pooling.
\newblock In {\em International Conference on Learning Representations}, 2021.

\bibitem{bianchi2020hierarchical}
Filippo~Maria Bianchi, Daniele Grattarola, Lorenzo Livi, and Cesare Alippi.
\newblock Hierarchical representation learning in graph neural networks with
  node decimation pooling.
\newblock {\em IEEE Transactions on Neural Networks and Learning Systems},
  2020.

\bibitem{borgwardt2005shortest}
Karsten~M Borgwardt and Hans-Peter Kriegel.
\newblock Shortest-path kernels on graphs.
\newblock In {\em International Conference on Data Mining}, pages 74--81, 2005.

\bibitem{brugere2018network}
Ivan Brugere, Brian Gallagher, and Tanya~Y Berger-Wolf.
\newblock Network structure inference, a survey: Motivations, methods, and
  applications.
\newblock {\em ACM Computing Surveys}, 51(2):1--39, 2018.

\bibitem{chen2019dagcn}
Fengwen Chen, Shirui Pan, Jing Jiang, Huan Huo, and Guodong Long.
\newblock Dagcn: dual attention graph convolutional networks.
\newblock In {\em International Joint Conference on Neural Networks}, pages
  1--8, 2019.

\bibitem{cubuk2019autoaugment}
Ekin~D Cubuk, Barret Zoph, Dandelion Mane, Vijay Vasudevan, and Quoc~V Le.
\newblock Autoaugment: Learning augmentation strategies from data.
\newblock In {\em IEEE Conference on Computer Vision and Pattern Recognition},
  pages 113--123, 2019.

\bibitem{dai2021hyperbolic}
Jindou Dai, Yuwei Wu, Zhi Gao, and Yunde Jia.
\newblock A hyperbolic-to-hyperbolic graph convolutional network.
\newblock In {\em {IEEE/CVF} Conference on Computer Vision and Pattern
  Recognition}, pages 154--163, 2021.

\bibitem{di2020mutual}
Xinhan Di, Pengqian Yu, Rui Bu, and Mingchao Sun.
\newblock Mutual information maximization in graph neural networks.
\newblock In {\em International Joint Conference on Neural Networks}, pages
  1--7. {IEEE}, 2020.

\bibitem{gilmer2017neural}
Justin Gilmer, Samuel~S Schoenholz, Patrick~F Riley, Oriol Vinyals, and
  George~E Dahl.
\newblock Neural message passing for quantum chemistry.
\newblock In {\em International Conference on Machine Learning}, pages
  1263--1272, 2017.

\bibitem{gomez2011uncovering}
M Gomez~Rodriguez, D Balduzzi, B Sch{\"o}lkopf, Getoor~T Scheffer, et~al.
\newblock Uncovering the temporal dynamics of diffusion networks.
\newblock In {\em International Conference on Machine Learning}, pages
  561--568, 2011.

\bibitem{hamilton2017inductive}
William~L Hamilton, Rex Ying, and Jure Leskovec.
\newblock Inductive representation learning on large graphs.
\newblock In {\em International Conference on Neural Information Processing
  Systems}, pages 1025--1035, 2017.

\bibitem{hassani2020contrastive}
Kaveh Hassani and Amir~Hosein Khasahmadi.
\newblock Contrastive multi-view representation learning on graphs.
\newblock In {\em International Conference on Machine Learning}, pages
  4116--4126, 2020.

\bibitem{hjelm2018learning}
R~Devon Hjelm, Alex Fedorov, Samuel Lavoie-Marchildon, Karan Grewal, Phil
  Bachman, Adam Trischler, and Yoshua Bengio.
\newblock Learning deep representations by mutual information estimation and
  maximization.
\newblock In {\em International Conference on Learning Representations}, 2019.

\bibitem{khosla2020supervised}
Prannay Khosla, Piotr Teterwak, Chen Wang, Aaron Sarna, Yonglong Tian, Phillip
  Isola, Aaron Maschinot, Ce Liu, and Dilip Krishnan.
\newblock Supervised contrastive learning.
\newblock In {\em Neural Information Processing Systems}, volume~33, 2020.

\bibitem{GCNmodel}
Thomas~N. Kipf and Max Welling.
\newblock Semi-supervised classification with graph convolutional networks.
\newblock In {\em International Conference on Learning Representations}, 2017.

\bibitem{klicpera2019predict}
Johannes Klicpera, Aleksandar Bojchevski, and Stephan G{\"u}nnemann.
\newblock Predict then propagate: Combining neural networks with personalized
  pagerank for classification on graphs.
\newblock In {\em International Conference on Learning Representations}, 2019.

\bibitem{kolouri2021wasserstein}
Soheil Kolouri, Navid Naderializadeh, Gustavo~K Rohde, and Heiko Hoffmann.
\newblock Wasserstein embedding for graph learning.
\newblock In {\em International Conference on Learning Representations}, 2021.

\bibitem{lee2019self}
Junhyun Lee, Inyeop Lee, and Jaewoo Kang.
\newblock Self-attention graph pooling.
\newblock In {\em International Conference on Machine Learning}, pages
  3734--3743, 2019.

\bibitem{li2019semi}
Jia Li, Yu Rong, Hong Cheng, Helen Meng, Wenbing Huang, and Junzhou Huang.
\newblock Semi-supervised graph classification: A hierarchical graph
  perspective.
\newblock In {\em The World Wide Web Conference}, pages 972--982, 2019.

\bibitem{li2020graph}
Maosen Li, Siheng Chen, Ya Zhang, and Ivor Tsang.
\newblock Graph cross networks with vertex infomax pooling.
\newblock In {\em Neural Information Processing Systems}, volume~33, pages
  14093--14105, 2020.

\bibitem{li2021braingnn}
Xiaoxiao Li, Yuan Zhou, Nicha Dvornek, Muhan Zhang, Siyuan Gao, Juntang Zhuang,
  Dustin Scheinost, Lawrence~H Staib, Pamela Ventola, and James~S Duncan.
\newblock Braingnn: Interpretable brain graph neural network for fmri analysis.
\newblock {\em Medical Image Analysis}, 74:102233, 2021.

\bibitem{liu2020tensor}
Xien Liu, Xinxin You, Xiao Zhang, Ji Wu, and Ping Lv.
\newblock Tensor graph convolutional networks for text classification.
\newblock In {\em AAAI Conference on Artificial Intelligence}, pages
  8409--8416, 2020.

\bibitem{liu2021self}
Xiao Liu, Fanjin Zhang, Zhenyu Hou, Li Mian, Zhaoyu Wang, Jing Zhang, and Jie
  Tang.
\newblock Self-supervised learning: Generative or contrastive.
\newblock {\em IEEE Transactions on Knowledge and Data Engineering}, 2021.

\bibitem{ma2019graph}
Yao Ma, Suhang Wang, Charu~C Aggarwal, and Jiliang Tang.
\newblock Graph convolutional networks with eigenpooling.
\newblock In {\em International Conference on Knowledge Discovery \& Data
  Mining}, pages 723--731, 2019.

\bibitem{narayanan2017graph2vec}
Annamalai Narayanan, Mahinthan Chandramohan, Rajasekar Venkatesan, Lihui Chen,
  Yang Liu, and Shantanu Jaiswal.
\newblock graph2vec: Learning distributed representations of graphs.
\newblock {\em CoRR}, abs/1707.05005, 2017.

\bibitem{nezhadarya2020adaptive}
Ehsan Nezhadarya, Ehsan Taghavi, Ryan Razani, Bingbing Liu, and Jun Luo.
\newblock Adaptive hierarchical down-sampling for point cloud classification.
\newblock In {\em {IEEE/CVF} Conference on Computer Vision and Pattern
  Recognition}, pages 12956--12964, 2020.

\bibitem{parisot2018disease}
Sarah Parisot, Sofia~Ira Ktena, Enzo Ferrante, Matthew Lee, Ricardo Guerrero,
  Ben Glocker, and Daniel Rueckert.
\newblock Disease prediction using graph convolutional networks: application to
  autism spectrum disorder and alzheimer’s disease.
\newblock {\em Medical Image Analysis}, 48:117--130, 2018.

\bibitem{peng2020graph}
Zhen Peng, Wenbing Huang, Minnan Luo, Qinghua Zheng, Yu Rong, Tingyang Xu, and
  Junzhou Huang.
\newblock Graph representation learning via graphical mutual information
  maximization.
\newblock In {\em The Web Conference}, pages 259--270, 2020.

\bibitem{qiu2020gcc}
Jiezhong Qiu, Qibin Chen, Yuxiao Dong, Jing Zhang, Hongxia Yang, Ming Ding,
  Kuansan Wang, and Jie Tang.
\newblock Gcc: Graph contrastive coding for graph neural network pre-training.
\newblock In {\em International Conference on Knowledge Discovery \& Data
  Mining}, pages 1150--1160, 2020.

\bibitem{scarselli2008graph}
Franco Scarselli, Marco Gori, Ah~Chung Tsoi, Markus Hagenbuchner, and Gabriele
  Monfardini.
\newblock The graph neural network model.
\newblock {\em IEEE transactions on neural networks}, 20(1):61--80, 2008.

\bibitem{shervashidze2011weisfeiler}
Nino Shervashidze, Pascal Schweitzer, Erik~Jan Van~Leeuwen, Kurt Mehlhorn, and
  Karsten~M Borgwardt.
\newblock Weisfeiler-lehman graph kernels.
\newblock {\em Journal of Machine Learning Research}, 12(9), 2011.

\bibitem{shervashidze2009efficient}
Nino Shervashidze, SVN Vishwanathan, Tobias Petri, Kurt Mehlhorn, and Karsten
  Borgwardt.
\newblock Efficient graphlet kernels for large graph comparison.
\newblock In {\em Artificial intelligence and statistics}, pages 488--495,
  2009.

\bibitem{shi2019diffusion}
Yong Shi, Minglong Lei, Hong Yang, and Lingfeng Niu.
\newblock Diffusion network embedding.
\newblock {\em Pattern Recognition}, 88:518--531, 2019.

\bibitem{shui2020heterogeneous}
Zeren Shui and George Karypis.
\newblock Heterogeneous molecular graph neural networks for predicting molecule
  properties.
\newblock In {\em IEEE International Conference on Data Mining}, pages
  492--500, 2020.

\bibitem{sugiyama2015halting}
Mahito Sugiyama and Karsten Borgwardt.
\newblock Halting in random walk kernels.
\newblock In {\em Neural Information Processing Systems}, volume~28, pages
  1639--1647, 2015.

\bibitem{sun2020infograph}
Fan-Yun Sun, Jordan Hoffmann, Vikas Verma, and Jian Tang.
\newblock Infograph: Unsupervised and semi-supervised graph-level
  representation learning via mutual information maximization.
\newblock In {\em International Conference on Learning Representations}, 2020.

\bibitem{sun2021sugar}
Qingyun Sun, Jianxin Li, Hao Peng, Jia Wu, Yuanxing Ning, Philip~S Yu, and
  Lifang He.
\newblock Sugar: Subgraph neural network with reinforcement pooling and
  self-supervised mutual information mechanism.
\newblock In {\em {WWW}: Web Conference}, pages 2081--2091, 2021.

\bibitem{velivckovic2018graph}
Petar Veli{\v{c}}kovi{\'c}, Guillem Cucurull, Arantxa Casanova, Adriana Romero,
  Pietro Lio, and Yoshua Bengio.
\newblock Graph attention networks.
\newblock In {\em International Conference on Learning Representations}, 2018.

\bibitem{velickovic2019deep}
Petar Velickovic, William Fedus, William~L Hamilton, Pietro Li{\`o}, Yoshua
  Bengio, and R~Devon Hjelm.
\newblock Deep graph infomax.
\newblock In {\em International Conference on Learning Representations}, 2019.

\bibitem{vignac2020building}
Clement Vignac, Andreas Loukas, and Pascal Frossard.
\newblock Building powerful and equivariant graph neural networks with
  structural message-passing.
\newblock In {\em Neural Information Processing Systems}, 2020.

\bibitem{vishwanathan2010graph}
S~Vichy~N Vishwanathan, Nicol~N Schraudolph, Risi Kondor, and Karsten~M
  Borgwardt.
\newblock Graph kernels.
\newblock {\em Journal of Machine Learning Research}, 11:1201--1242, 2010.

\bibitem{wang2019dynamic}
Yue Wang, Yongbin Sun, Ziwei Liu, Sanjay~E Sarma, Michael~M Bronstein, and
  Justin~M Solomon.
\newblock Dynamic graph cnn for learning on point clouds.
\newblock {\em ACM Transactions On Graphics}, 38(5):1--12, 2019.

\bibitem{xinyi2018capsule}
Zhang Xinyi and Lihui Chen.
\newblock Capsule graph neural network.
\newblock In {\em International conference on learning representations}, 2018.

\bibitem{xu2018powerful}
Keyulu Xu, Weihua Hu, Jure Leskovec, and Stefanie Jegelka.
\newblock How powerful are graph neural networks?
\newblock In {\em International Conference on Learning Representations}, 2019.

\bibitem{xu2018representation}
Keyulu Xu, Chengtao Li, Yonglong Tian, Tomohiro Sonobe, Ken-ichi Kawarabayashi,
  and Stefanie Jegelka.
\newblock Representation learning on graphs with jumping knowledge networks.
\newblock In {\em International Conference on Machine Learning}, pages
  5453--5462, 2018.

\bibitem{xuan2019subgraph}
Qi Xuan, Jinhuan Wang, Minghao Zhao, Junkun Yuan, Chenbo Fu, Zhongyuan Ruan,
  and Guanrong Chen.
\newblock Subgraph networks with application to structural feature space
  expansion.
\newblock {\em IEEE Transactions on Knowledge and Data Engineering}, 2019.

\bibitem{yanardag2015deep}
Pinar Yanardag and SVN Vishwanathan.
\newblock Deep graph kernels.
\newblock In {\em International Conference on Knowledge Discovery and Data
  Mining}, pages 1365--1374, 2015.

\bibitem{yang2021selfsagcn}
Xu Yang, Cheng Deng, Zhiyuan Dang, Kun Wei, and Junchi Yan.
\newblock Selfsagcn: Self-supervised semantic alignment for graph convolution
  network.
\newblock In {\em {IEEE/CVF} Conference on Computer Vision and Pattern
  Recognition}, pages 16775--16784, 2021.

\bibitem{yao2019graph}
Liang Yao, Chengsheng Mao, and Yuan Luo.
\newblock Graph convolutional networks for text classification.
\newblock In {\em AAAI conference on artificial intelligence}, pages
  7370--7377, 2019.

\bibitem{ying2018hierarchical}
Rex Ying, Jiaxuan You, Christopher Morris, Xiang Ren, William~L Hamilton, and
  Jure Leskovec.
\newblock Hierarchical graph representation learning with differentiable
  pooling.
\newblock In {\em Neural Information Processing Systems}, pages 4805--4815,
  2018.

\bibitem{you2020graph}
Yuning You, Tianlong Chen, Yongduo Sui, Ting Chen, Zhangyang Wang, and Yang
  Shen.
\newblock Graph contrastive learning with augmentations.
\newblock In {\em Neural Information Processing Systems}, volume~33, pages
  5812--5823, 2020.

\bibitem{zhang2018end}
Muhan Zhang, Zhicheng Cui, Marion Neumann, and Yixin Chen.
\newblock An end-to-end deep learning architecture for graph classification.
\newblock In {\em AAAI Conference on Artificial Intelligence}, 2018.

\bibitem{zhang2019hierarchical}
Zhen Zhang, Jiajun Bu, Martin Ester, Jianfeng Zhang, Chengwei Yao, Zhi Yu, and
  Can Wang.
\newblock Hierarchical graph pooling with structure learning.
\newblock {\em CoRR}, abs/1911.05954, 2019.

\bibitem{zhang2018retgk}
Zhen Zhang, Mianzhi Wang, Yijian Xiang, Yan Huang, and Arye Nehorai.
\newblock Retgk: graph kernels based on return probabilities of random walks.
\newblock In {\em Neural Information Processing Systems}, pages 3968--3978,
  2018.

\bibitem{zhu2020gssnn}
Shichao Zhu, Lewei Zhou, Shirui Pan, Chuan Zhou, Guiying Yan, and Bin Wang.
\newblock Gssnn: graph smoothing splines neural networks.
\newblock In {\em AAAI Conference on Artificial Intelligence}, pages
  7007--7014, 2020.

\bibitem{zhu2021graph}
Yanqiao Zhu, Yichen Xu, Feng Yu, Qiang Liu, Shu Wu, and Liang Wang.
\newblock Graph contrastive learning with adaptive augmentation.
\newblock In {\em Web Conference}, pages 2069--2080, 2021.

\end{thebibliography}
}

\end{document}